\documentclass[pmlr,onecolumn]{jmlr} 
 
\usepackage{booktabs}
\usepackage[load-configurations=version-1]{siunitx} 
\theorembodyfont{\upshape}
\theoremheaderfont{\scshape}
\theorempostheader{:}
\theoremsep{\newline}

\jmlrvolume{ML4H Extended Abstract Arxiv Index}
\jmlryear{2020}
\jmlrsubmitted{2020}
\jmlrpublished{}
\jmlrworkshop{Machine Learning for Health (ML4H) 2020} 

\title[Forecasting Emergency Department Capacity Constraints for COVID Isolation Beds]{Forecasting Emergency Department Capacity Constraints for COVID Isolation Beds}

\author{
\Name{Erik Drysdale} \Email{erik.drysdale@sickkids.ca}\\
\addr The Hospital for Sick Children, Toronto, Ontario \\
\AND
\Name{Dr. Devin Singh} \Email{devin.singh@sickkids.ca}\\
\addr The Hospital for Sick Children, Toronto, Ontario \\
\AND
\Name{Dr. Anna Goldenberg} \Email{anna.goldenberg@sickkids.ca} \\
\addr The Hospital for Sick Children, Toronto, Ontario 
}

\begin{document}

\maketitle

\begin{abstract}
Predicting patient volumes in a hospital setting is a well-studied application of time series forecasting. Existing tools usually make forecasts at the daily or weekly level to assist in planning for staffing requirements. Prompted by new COVID-related capacity constraints placed on our pediatric hospital's emergency department, we developed an hourly forecasting tool to make predictions over a 24 hour window. These forecasts would give our hospital sufficient time to be able to martial resources towards expanding capacity and augmenting staff (e.g. transforming wards or bringing in physicians on call). Using Gaussian Process Regressions (GPRs), we obtain strong performance for both point predictions (average R-squared: 82\%) as well as classification accuracy when predicting the ordinal tiers of our hospital's capacity (average precision/recall: 82\%/74\%). Compared to traditional regression approaches, GPRs not only obtain consistently higher performance, but are also robust to the dataset shifts that have occurred throughout 2020. Hospital stakeholders are encouraged by the strength of our results, and we are currently working on moving our tool to a real-time setting with the goal of augmenting the capabilities of our healthcare workers. 
\end{abstract}

\begin{keywords}
Emergency Department; Gaussian Process Regression; Forecasting
\end{keywords}

\section{Introduction \label{sec:intro}}

The emergency department's (ED) census, the total number of patients admitted or waiting to be admitted, contains recurring patterns such as seasonality and hourly trends \citep{Flottemesch2007,nas2019}. Though classical time series approaches have been shown to work relatively well for predicting patient volumes \citep{Tandberg1994}, there is still scope for improvement. Most existing census forecasts tend to use daily or weekly time frames, as these allow for staffing decisions to be made in advance \citep{Rotstein1997,Batal2001,Boyle2008,Yeung2009,Kam2010,Calegari2016,Capan2016,Guan2019}. Hourly forecasting models have also been developed, but these almost all use ARIMA-type models \citep{Morzuch2006, Schweigler2009, Whitt2019} or generalized linear models \citep{McCarthy2008,Asheim2019}, and only occasionally neural networks \citep{Choudhury2019}. 

Though the existing literature is large, our research provides the following contributions. First, we show that parametric models are subject to a performance collapse when COVID causes a dataset shift to our  time series. Second, a small training set and an implicit transfer learning paradigm is demonstrated to obtain the best performance. Third, GPRs obtain high prediction accuracy at the hourly frequency for our hospital's pediatric census. GPRs are a natural fit for our modelling task as they 1) can be easily calibrated through their probabilistic inference, and 2) allow for kernels to be constructed that match our census' statistical patterns. Lastly, we show how a machine learning (ML) tool can be developed and evaluated to address a clinical use case in the COVID-era.

\section{Data}

All ED visits from June 1st, 2018 to August 30th, 2020 were extracted from our hospital's Epic database, amounting to 159,683 unique patient visits. Patient arrival and discharge time were used to calculate the maximum census of ED patients in any one hour. Additional demographic, physiological, and hospital information was processed. Categorical features, like sex, were calculating as an hourly share, and continuous features, like blood pressure, as an hourly average. Appendix \ref{apd:first} provides a full description of all preprocessing steps.

Since the onset of COVID, our pediatric hospital's ED has operated under strict capacity limits with a total of 47 isolation beds. Any patient showing a possible COVID symptom must be placed in one of these areas. Since common symptoms like a runny nose or fever are sufficient for a patient to be placed in an isolation room, a large fraction of the census use up these scarce resources. 

\begin{figure}[htbp]
  \centering
  \includegraphics[width=0.6\linewidth]{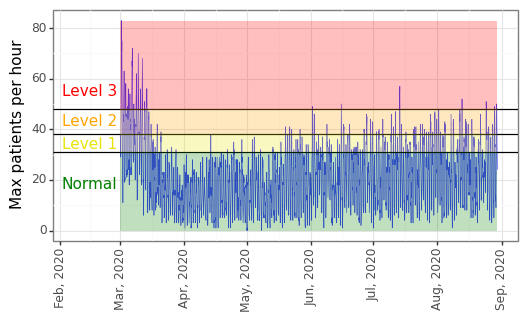}
  \caption{Hospital capacity constraints at the hourly level \label{fig:capacity}}
\end{figure}

In our analysis we assumed that 100\% of patients required an isolation bed. Though this will not be the case in practice, we lacked historical data to determine this share in any one hour and therefore opted to be conservative in the estimation of our ED's capacity constraints. Figure \ref{fig:capacity} shows the trend in the maximum number of hourly patients in our ED since March 2020. As the ED census moves from the normal state ($\leq 30$) into escalation level 1 (31-37), level 2 (38-47), or level 3 ($\geq 48$), a variety of prescribed actions are carried out to free up additional hospital space and martial more resources. Under our 100\% isolation assumption, if the escalation level is 3, then our ED will be unable to admit any new patients. This represents a significant harm to our patients, community, and healthcare workers. Our hospital will benefit from an early warning system of impending escalation level changes.

\section{Modelling and evaluation}

Starting on January 1st, 2020 models were required to make a rolling forecast every hour, for a 24 hour window, with retraining occurring at midnight. For example, the training set available to a model for February 3rd, 2020 would include all hourly data from June 1st, 2018 until midnight on February 2nd, 2020. The GPR was benchmarked against a Lasso model provided with autoregressive (AR) features \citep{Rasmussen2005,tibs1996}, as well as one weighted by local observations (see Appendix \ref{apd:second} for a full list of features). The AR-Lasso used features that were of the current hour, as well as those of the previous ten hours. The level of regularization was selected to bound the number of false positives to 10\% \citep{Drysdale2019}. For the locally weighted AR-Lasso, a validation set was created by taking the week of data before the forecasting day, and selecting a length scale for the radial basis function kernel which minimized that week's mean square error (MSE).

The GPR was fit with the \texttt{GPyTorch} package \citep{Gardner2018} using an Adam optimizer (learning rate = 0.1), and was trained until the change in marginal log-likelihood was less that 1e-3. At the time of nightly retraining, the previous day's weights were loaded (the hyperparameters of the various kernels). The GPR kernel combined multiple kernel types (linear, cosine, and RBF) to capture cyclical trends, up-weight recent observations, and match hours that had similar features (see Appendix \ref{apd:third} for a complete specification of the kernel). The feature set used by the GPR comprised of the time series for previous values of the regression label (the maximum number of patients in a given hour), the number of arrivals and discharges, the hour of the day, the composition of patients by acuity score and arrival method, as well as the number of physicians in the ED. These features were chosen as they were amenable for real-time extraction and consistent with clinical expertise.  For each regression method, a separate model was trained for a given forecasting lead. 

All one-day-ahead forecasting results were analyzed from March 1st to August 31st. Point estimates were evaluated by calculating the daily R-squared (R2) for each forecasting lead. Due to the importance of our hospital's capacity constraints and prescribed action required for different escalation levels (see Figure \ref{fig:capacity}), we evaluated the performance of our model by how well it predicted positive changes to the current escalation level. For example, if the ED census was 32 at the current hour (level 1), and the 6-hour forecasting lead predicted 49 patients (level 3), this would be a predicted positive change of two levels. Predicted positive changes are therefore forecasts where the level of patients corresponds to a higher escalation level than what is seen at the current hour. By treating the differences in the levels as different classes, precision and recall could be calculated. Specifically, there is a class for positive level changes of one, two, or three, relative to the current level. A less stringent class is that of any positive level change (i.e. one, two, or three).

\section{Results}

\begin{figure}[!h]
\floatconts
  {fig:subfigex}
  {\caption{Variance explained by model and training method}}
  {%
    \subfigure[AR models vs the GPR]{\label{fig:para_vs_gp}%
      \includegraphics[width=0.45\linewidth]{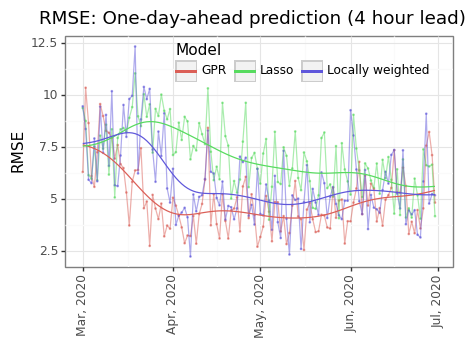}}%
    \qquad
    \subfigure[Iterative vs no-pretraining]{\label{fig:train_iter}%
      \includegraphics[width=0.40\linewidth]{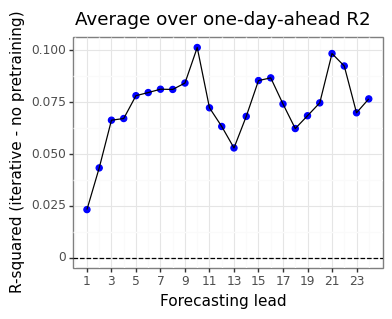}}
  }
\end{figure}

The AR models showed a fragile one-day-ahead forecasting performance. Figure \ref{fig:para_vs_gp} demonstrates this phenomenon for the 4-hour lead (other leads showed the same dynamic). As patient volumes fell in March, the AR-Lasso actually saw an increase in its root MSE (RMSE), despite the smaller variation in the label, for two reasons. First, the expected level of patients was systematically too high, and second, the relative ranking deteriorated as measured by the concordance index. Even after the RMSE of the AR-Lasso began to cover, its performance never converged to the level of the GPR. Although the locally weighted model performed better than the AR-Lasso, it still lagged behind the GPR.

Due to the  non-stationary nature of the ED census, the optimal size of training data for the GPR was also considered. We found that using 1-2 months of training data yielded a much lower performance ($\approx 65\%$ R2) compared to 1-2 weeks ($\approx 80\%$ R2) or even 1-3 days ($\approx 85\%$ R2). Excluding the 1-hour lead, the performance for leads 2-24 was similar, on average. Even though a short time horizon is clearly beneficial for adapting to dataset shifts, the magnitude of the performance obtainable using only a few training days was surprising. However an examination of the relative performance differences from iterative retraining (using the previous day's hyperparameters) compared to fitting the GPR from scratch revealed that the latter, \textit{de novo} model training, would result in a significant performance decline ($\approx 7.5\%$ R2) as Figure \ref{fig:train_iter} shows.

\begin{figure}[!h]
\vspace{-0.1cm}
\floatconts
  {fig:subfigex}
  {\caption{Precision for escalation levels}}
  {%
    \subfigure[Precision by escalation levels; CIs based on exact binomial distribution quantiles]{\label{fig:precision}%
      \includegraphics[width=0.45\linewidth]{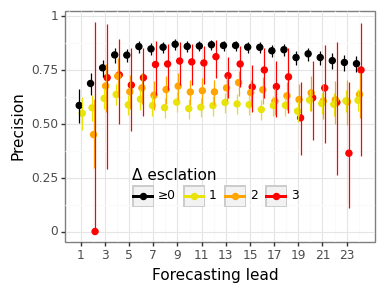}}%
    \qquad
    \subfigure[Precision/Recall curves ($\Delta>0$ escalation)]{\label{fig:pr_curve}%
      \includegraphics[width=0.45\linewidth]{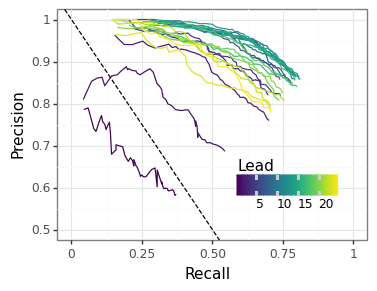}}
    \qquad
    \subfigure[Number of hourly escalation levels in 2020 by lead]{\label{fig:nlbls}
     \includegraphics[width=0.70\linewidth]{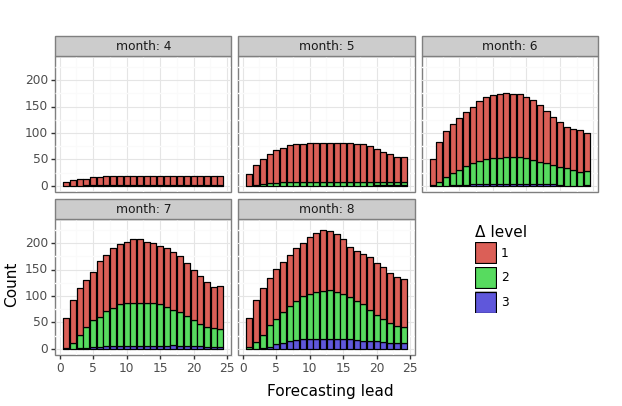}}
  }
\end{figure}

Using point estimates of the GPR, the model was able to accurately predict a positive change in the escalation level (e.g. going from a normal state ($\leq 30$ patients), to level 2 ($38-47$ patients)) as measured by both precision and recall. Figure \ref{fig:precision} shows that the average precision was between 50-75\% for predicting the \textit{exact} (positive) change in escalation level, and less stringently, 80\% for any positive change. Because the GPR provides a probabilistic output, a trade-off between precision and recall can be obtained by using different percentiles of the output. Figure \ref{fig:pr_curve} shows that sensitivity can be traded off between 20-70\% depending on a tolerance for false positives for most forecasting leads. Conditional on a given recall, precision precision shows a U-shaped relationship for the forecasting lead. This is due to label balance peaking at the 12 hour lead, as figure \ref{fig:nlbls} shows.

\section{Conclusion}

The GPR provides sufficient flexibility to handle non-stationary ED census data in the COVID-era and obtains a strong performance, especially when compared to more brittle parametric models. Using a small number of training days, having nightly retraining, and loading pre-existing weights obtains the best performance. We have shown how model performance can be explicitly linked to a clinical application in our hospital, namely: predicting escalation levels to allow for early operational decisions. Our hospital stakeholders are impressed with the results, and are eager to test our tool in a real-time setting. These forecasts show clear potential to augment and enhance the capabilities of healthcare workers.

\newpage

\acks{Thanks to Dr. Jason Fischer and Dr. Tania Principi for their clinical insight and support.}

\bibliography{jmlr-sample}

\appendix


\section{Data preprocessing \label{apd:first}}

After the raw data was extracted from our hospital's database with Epic Reporting Workbench, the data underwent a series of transformations to account for missing values, simple feature transformations, and to remove obvious outliers. A small number of visits had no arrival time (24) or no length of stay (74) and were removed from the dataset (out of 159,707 rows).

\begin{enumerate}
    \item Distance to hospital: the patient's postal code was converted into a latitude/longitude measure and then compared to our hospital's coordinates using the Haversine distance. The distances were then logged and binned into approximate sextiles: 0-1, 1-2, 2-3, 3-4, 4-10, and 10+ logged kilometers. 
    \item Language: Missing values were encoded as ``missing''. Any language that was recorded as occurring in less than 0.1\% of records was set to ``other''.
    \item Respiratory and pulse values outside of the range 0-80 and 20-220, respectively, were set as missing.
    \item Blood pressure was split into a systolic and diastolic component.
    \item A dummy was recorded as to whether the patient had been discharged from the ED previously within 72 hours. 
    \item Sex values left blank or reported as ``unknown'' were randomly assigned (only 9 instances).
    \item All arrival methods were mapped to one of 7 categories: Air, Ambulance, Car, Other, Transfer, Walk.
    \item The acuity score (CTAS) was set to ``missing'' when a patient had no score listed.
    \item Length of stay (LOS): Our database recorded two LOS measures, and the maximum of the two was used to calculate the discharge time. If the maximum LOS was  equal to zero, the value was set to one minute.
    \item Discharge time: Applied the calculated LOS (see above) to the recorded arrival time.
    \item Lab orders: Any lab orders within a given hour were counted by their type. Orders that occurred less than 1\% of the time were set to ``other''. 
    \item Diagnostic imaging (DIs): DIs were processed in the same way as lab orders.
    \item Physician team: The number of physicians listed in a patient's recorded were counted and stored in a list allowing for an hourly average and a unique count.
\end{enumerate}

\section{Final feature set \label{apd:second}}

A total of 163 features were used by the Lasso models, and were a mix of continuous (CONT) or one-hot-encoded (OHE) variables. Each feature had lags generated for 0 to 10 hours ago. Unless otherwise noted, each feature had a version that calculated its value for arrivals versus discharges that hour. All values were normalized using the mean and standard deviations calculated on the training set.

\begin{enumerate}
    \item Language: 26 categories (OHE)
    \item Arrival method: 15 categories (OHE)
    \item CTAS: 6 levels (OHE)
    \item Distance to the hospital: 6 levels (OHE)
    \item Sex: 2 levels (OHE)
    \item 72 Return indicator: 1 level (OHE)
    \item Average number of patient medications (CONT)
    \item Pulse, Resp, Weight, Diastolic, Systolic, Temperature, Age (CONT)
    \item Count of arrivals/discharges (CONT)
    \item Average number of MDs (CONT)
    \item Unique number of MDs in the past 10 hours; no arrival-discharge pair
    \item Laboratory counts: 21 measures (CONT); no arrival-discharge pair
    \item DI counts: 7 measures (CONT); no arrival-discharge pair
    \item Census: both the maximum and the variation in a given hour; no arrival-discharge pair
\end{enumerate}

\section{GPR Kernel \label{apd:third}}

\newcommand{\bx}{\textbf{x}}
\newcommand{\bt}{\textbf{t}}

The vector $\bx$ includes all features described in Appendix \ref{apd:second}, where $\bt$ is a time index (number of hours, indexed to start of training set), and $\bx_i$ is the subset of features that have been grouped for the GPR. All features and the response were normalized. The GPR-specific features were as follows: the hourly census maximum and its variance, the count of arrivals and discharges, the hour of the day, the composition of patients by acuity score and arrival method, as well as the average number of physicians in the ED. This specific feature set was chosen because 1) these variables are easy to capture to in a real-time environment, 2) they showed (relatively) strong performance for univariate GPR models, and 3) they were intuitive to our clinical leads.

\begin{align*}
    f(\bx)  &\sim \text{GP}\big(m(\bx),K(\bx,\bx')\big)  \\
    K_1(\bx,\bx') &= K^{\textbf{Cos}}_1(\bt, \bt') +  K^{\textbf{Lin}}_1(\bt, \bt') + \\
    &\hspace{7mm}\sum_i K_{1i}^{\text{RBF}}(\bx_i, \bx_i')  \\
    K_2(\bx,\bx') &= K^{\textbf{Cos}}_2(\bt, \bt') \cdot  K^{\textbf{Lin}}_2(\bt, \bt') +  \\
    &\hspace{7mm} \sum_i K_{2i}^{\text{RBF}}(\bx_i, \bx_i') \cdot K^{\textbf{Lin}}_{2i}(\bt, \bt') \\
    K(\bx,\bx') &= K_1 + K_2 + K_1 \cdot K_2 \\
    m(\bx) &= \textbf{0} \\
\end{align*}

\end{document}